\documentclass[conference]{IEEEtran}
\IEEEoverridecommandlockouts
\usepackage{amsmath}

\DeclareMathOperator*{\argmin}{arg\,min}
\usepackage{amssymb,amsfonts}
\usepackage{graphicx}
\usepackage{textcomp}
\usepackage{xcolor}
\usepackage{algorithm}
\usepackage[noend]{algpseudocode}
\usepackage[flushleft]{threeparttable}
\usepackage{array}

\newcommand{\pluseq}{\mathrel{+}=}
\newcommand{\etal}{\emph{et al. }}

\begin{document}

\title{Can't Boil This Frog: Robustness of Online-Trained Autoencoder-Based Anomaly Detectors to Adversarial Poisoning Attacks}


\author{\IEEEauthorblockN{Moshe Kravchik}
\IEEEauthorblockA{\textit{Dept. of Software and Information Systems Engineering} \\
\textit{Ben-Gurion University of the Negev}\\
Beer-Sheva, Israel\\
moshekr@post.bgu.ac.il}
\and
\IEEEauthorblockN{Asaf Shabtai}
\IEEEauthorblockA{\textit{Dept. of Software and Information Systems Engineering } \\
\textit{Ben-Gurion University of the Negev}\\
Beer-Sheva, Israel\\
shabtaia@bgu.ac.il}
}
\maketitle

\begin{abstract}
In recent years, a variety of effective neural network-based methods for anomaly and cyber attack detection in industrial control systems (ICSs) have been demonstrated in the literature. 
Given their successful implementation and widespread use, there is a need to study adversarial attacks on such detection methods to better protect the systems that depend upon them. 
The extensive research performed on adversarial attacks on image and malware classification has little relevance to the physical system state prediction domain, which most of the ICS attack detection systems belong to.
Moreover, such detection systems are typically retrained using new data collected from the monitored system, thus the threat of adversarial data poisoning is significant, however this threat has not yet been addressed by the research community.
In this paper, we present the first study focused on poisoning attacks on online-trained autoencoder-based attack detectors.
We propose two algorithms for generating poison samples, an interpolation-based algorithm and a back-gradient optimization-based algorithm, which we evaluate on both synthetic and real-world ICS data.
We demonstrate that the proposed algorithms can generate poison samples that cause the target attack to go undetected by the autoencoder detector, however the ability to poison the detector is limited to a small set of attack types and magnitudes.
When the poison-generating algorithms are applied to the popular SWaT dataset, we show that the autoencoder detector trained on the physical system state data is resilient to poisoning in the face of all ten of the relevant attacks in the dataset.
This finding suggests that neural network-based attack detectors used in the cyber-physical domain are more robust to poisoning than in other problem domains, such as malware detection and image processing.
\end{abstract}


\section{Introduction}\label{sec:introduction}
Neural network-based anomaly and attack detection methods have attracted significant attention in recent years.
The ability of neural networks (NNs) to accurately model complex multivariate data has contributed to their use in detectors in various areas ranging from medical diagnostics to malware detection and cyber-physical systems monitoring.
An attack detection system's effectiveness depends heavily on its robustness to attacks that target the detection system itself.
In the context of NNs, such attacks are known as adversarial data attacks.
Adversarial attacks have been a major focus of the NN research community, primarily in the image classification (\cite{suciu2018does, shafahi2018poison}), malware detection (\cite{suciu2018does,rosenberg2018generic}), and network intrusion detection (\cite{rubinstein2009antidote, madani2018robustness}) domains.

This research focuses on NN-based anomaly and cyber attack detectors in industrial control systems (ICSs) which are a subclass of cyber-physical systems (CPSs).
ICSs are central to many important areas of industry, energy production, and critical infrastructure.
The security and safety of ICSs are therefore of the utmost importance.
While a number of recent studies have proposed using NN-based detectors to improve ICS security, the resilience of such detectors to adversarial attacks has received little attention.

Adversarial attacks on NNs can be broadly divided into poisoning and evasion attacks.
Both kinds of attacks use maliciously crafted data to achieve their goals.
\textbf{Evasion} attacks aim to craft test data samples that will evade detection while still producing the desired adversarial effect (e.g., service disruption).
Recently, three studies examining adversarial attacks on NN ICS detectors were published, all of which were dedicated to evasion attacks (\cite{erba2019real,kravchik2019efficient,zizzo2019intrusion}) performed using different threat models.

On the other hand, \textbf{poisoning} attacks attempt to introduce adversarial data during the model's training.
This data influences the model in such a way that the target attack remains undetected at test time.
However, the model's detection behavior for all other inputs is maintained, ensuring that no suspicion of an attack is raised.
The importance of poisoning attack research has increased in the light of the popularity of ICS monitoring systems' online training mode.
With online training, the model is periodically trained with new data collected from the protected system to accommodate for the concept drift.
The fact that this retraining provides the adversary with the opportunity to poison the model underscores the need to study poisoning attacks on ICS anomaly and attack detectors.
However, this task is challenging.
According to \cite{jagielski2018manipulating}, the impact of adversarially manipulated training data on the output of regression learning models (which are used in most NN ICS detectors) is not well understood.
Until now, the vast majority of poisoning research has dealt with classification problems, while regression learning poisoning has received less attention, with a focus on simpler machine learning algorithms, such as linear regression \cite{jagielski2018manipulating}.

To the best of our knowledge, this is the first study addressing poisoning attacks on deep semi-supervised NN detectors for physics-based multivariate time sequences, which are common in ICSs.
In this research, we investigate the following unique combination of issues related to anomaly and attack detection in ICSs.
\begin{enumerate}
    \item It targets a \textbf{semi-supervised machine learning detector}, while prior research largely examined adversarial attacks on supervised learning.
    \item It aims to poison a  \textbf{regression-based deep NN detector}, while prior research dealt mostly with classification problems.
    \item Our proposed solution is capable of identifying a \textbf{sequence} of \textit{series of vectors}, while previous studies considered single poisoning data points.
    \item Given the \textbf{online nature of the training}, the order of the poisoning inputs is relevant, an issue which has not been considered in previous research.
    \item Most NN ICS detectors process data in multiple overlapping time series, with the \textbf{same data serving as the target output for some predictions and the input for others}, magnifying the influence of the manipulation of a single time point.
\end{enumerate}

Our study aims to answer the following research questions.
\begin{enumerate}
    \item What algorithms can be used to generate poisoning input for a NN-based ICS anomaly detector operating in online training mode?
    \item How robust are the detectors proposed in \cite{taormina2018deep,kravchik2019efficient,erba2019real} to such poisoning attacks?
\end{enumerate}

The contributions of this paper are as follows:
\begin{itemize}
    \item We present the first study of poisoning attacks on online trained NN-based detectors for multivariate time series.
    \item We propose two algorithms for the generation of poisoning samples in such settings: an interpolation-based algorithm and a back-gradient optimization-based algorithm.
    \item We implement and validate both algorithms on synthetic data and evaluate the influence of various test parameters on the poisoning abilities of the algorithms.
    \item We apply the algorithms to an autoencoder-based detector for real-world ICS data and study the detector's robustness to poisoning attacks.
    \item The implementation of both algorithms and the evaluation tests code are open source and freely available.\footnote{The link will be provided after the review to preserve anonymity}
\end{itemize}

\section{Background}\label{sec:background}
In this section, we introduce the notation and provide background on ICSs, ICS anomaly detection, and poisoning attacks.

\subsection{Industrial control systems and anomaly detection}\label{sec:background_ics}
ICSs are comprised of network-connected  computers that monitor and control physical processes.
These computers obtain feedback about the monitored process from sensors and can influence the process using actuators, such as pumps, engines, and valves.
Typically, the sensors and actuators are connected to a local computing element, a programmable logic controller (PLC), which is a real-time specialized computer that runs a control loop supervising the physical process.
The PLCs, sensors, and actuators form a remote segment of the ICS network.
Other important ICS components reside in another network segment, the control segment, whose constituents typically include a supervisory control and data  acquisition (SCADA) workstation, a human-machine interface (HMI) machine, and a historian server.
The SCADA computer runs the software responsible for programming and controlling the PLC.
The HMI receives and displays the current state of the controlled process, and the historian keeps a record of all of the sensory and state data collected from the PLC.

In our research, we consider another component residing in the control segment, an anomaly and attack detector, which is sometimes also referred to as an intrusion detection system (IDS).
Its role is to analyze the system state and detect possible anomalies and attacks on the system.
For that purpose, the detector can use all available information, including network traffic and sensory data.

\begin{figure}[htpb]
\centering{\includegraphics[clip, trim=0cm 0cm 0cm 0cm, scale=0.4]{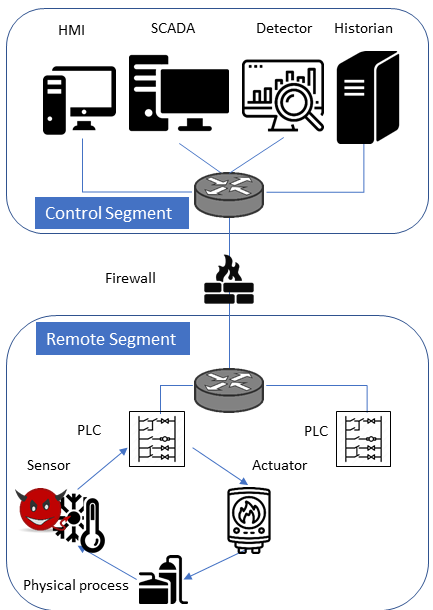}}
\caption{A schematic SCADA system diagram under the selected threat model.}
\label{fig:system}
\end{figure}

Various approaches for building such detectors were surveyed in \cite{mitchell2014survey}, \cite{humayed2017cyber}, and \cite{giraldo2017security}; of these, physics-based attack detectors are the most relevant to our research.
Such detectors are based on a fundamental idea that the behavior of the protected system is bound by immutable laws of physics and therefore can be modeled with sufficient precision.
The anomalous behavior is detected when the observed physical system state deviates from the model-based prediction.
In a recent paper, Giraldo \etal \cite{giraldo2018survey} reviewed close to fifty physics-based IDSs.

Our research focuses on detectors that use NNs to model the monitored process.
Recently, this approach has become very popular (e.g., \cite{kravchik2018detecting,taormina2018deep,lin2018tabor,taormina2018battle,erba2019real,kravchik2019efficient}), due to the ability of NNs to model complex multivariate systems with non-linear dependencies between the variables.
In the literature, various NN architectures have been used for anomaly and intrusion detection, including convolutional NNs, recurrent NNs, feed-forward NNs, and autoencoders \cite{goodfellow2016deep} (an extensive review was provided in \cite{chalapathy2019deep}).

While our research method is agnostic to the detector's architecture, we chose to evaluate our method using autoencoders due to their simplicity and popularity \cite{taormina2018deep,kravchik2019efficient,erba2019real}.
In the sections that follow, we introduce the notation and provide the necessary background on poisoning attacks.

\subsection{Notation}\label{sec:background_notation}
We follow the notation used in \cite{biggio2013security} and \cite{munoz2017towards}.
Unlike classification problems that distinguish between instance space and label space, our problem deals with a single feature space $\mathcal{Y}$.
In the context of ICS IDSs, the elements of this space are the sensor measurements and actuator states observed by the PLC and collected in by the historian.
These elements, $y_i \in \mathcal{R}^N$, are N-dimensional real-valued vectors representing the values of features (e.g., different sensors' readings) at time $i$.
In the general case, the goal of the model $\mathcal{M}$ is to predict the future feature values based on the past:
\begin{equation}
  \label{model_func}
  (\hat{y}_{h+n}, \hat{y}_{h+n+1}, \ldots, \hat{y}_{h+n+m}) = f(y_{n-1-l}, \ldots, y_{n-1}),
\end{equation}
where $y_i$ is a feature vector at time $i$, $\hat{y}_i$ is the estimation of the feature vector, $l$ and $m$ represent the input and output sequence length respectively, and $h$ is the prediction horizon.
In other words, the model uses historical values of the system state for the last $l$ time steps to predict the future values of the state for some $m$ time steps starting at the $h$-th time step from now.

For undercomplete autoencoders \cite{goodfellow2016deep}, the model learns a slightly different function - reconstructing the feature vector sequence from its compact code representation:
\begin{equation}
  \label{autoenc_func}
  \begin{array}{lcl}
  z_n = f_e(y_{n-l}, \ldots, y_{n}) \\
  (\hat{y}_{n-l}, \ldots, \hat{y}_{n}) = f_d(z_n),
  \end{array}
\end{equation}
where $z_n$ is a vector in code space $\mathcal{Z}$, $z_i \in \mathcal{R}^M$, and $f_e$ and $f_d$ are the functions learned respectively by the model's encoder and decoder parts.
An important property of the code vector $z_n$ is its size which must be less than the size of the input sequence, $M < N \cdot l$.
The learning is performed by minimizing an objective function $\mathcal{L}(\mathcal{D},w)$ on a given training set $\mathcal{D}_{tr}=\{y_i\}^n_{i=1}$, where $w$ are the model and learning parameters and hyperparameters.
This objective function $\mathcal{L}$  might include regularization, and hence we introduce $L(\mathcal{D}, w)$ to denote the loss function evaluated on the dataset $D$ using parameters $w$.

\subsection{Poisoning attacks}\label{sec:background_poisoning}
This research focuses on poisoning attacks.
In such attacks, the goal of the adversary is to influence the model by tampering with its training input.
There are two possible goals of such tampering: harming the detector's availability or its integrity.
In the context of regression-based anomaly detection, the aim of an availability attack is to cause the detector to generate numerous false alarms for valid data, while the aim of an integrity attack is to cause the detector to fail to detect an attack by predicting values close enough to the values of the features during the attack.

Previous work on regression learning poisoning focused on availability attacks~\cite{jagielski2018manipulating}.
Our research targets the detector's integrity: given an attack sequence $\mathcal{Y}_a = (y^1_a, \dots,  y^n_a)$, our goal is to produce a \textbf{sequence} of \textit{poisoning input sequences} $\mathcal{D}_p = (y^1_p, \dots, y^m_p)^k_{j=1}$ so that when trained with $\mathcal{D}_{tr} \cup \mathcal{D}_p$, the detector will not issue an alert upon encountering $\mathcal{Y}_{a}$ or issue false alerts for the untainted data.

The attacker's capabilities depend on his/her knowledge of the targeted system.
This can include knowledge on the training set $\mathcal{D}_{tr}$, the feature set $\mathcal{Y}$, the learning algorithm $\mathcal{M}$ and its hyperparameters, the learning objective function $\mathcal{L}$, and the learned parameters $w$ \cite{munoz2017towards}.
Thus, the attacker's knowledge can be described in terms of space $\Theta$, comprised of elements $\theta=(\mathcal{D}_{tr}, \mathcal{Y}, \mathcal{L}, w)$.
The white-box attack scenario in which the attacker knows all of the components of $\theta$ represents the worst case, and this is the scenario we are considering in this study.
Given an anomaly detection function $\mathcal{A}$($\mathcal{D}$,$w$) which returns the number of alerts for the given input set $\mathcal{D}$ and a model with trained weights $w$, the goal of our research is to find such $\mathcal{D}_p$ that the following holds true:
\begin{equation}
  \label{objective_func}
  \mathcal{A}(\mathcal{D}_{val} \cup \mathcal{Y}_a, w) = 0,
\end{equation}
where $\mathcal{D}_{val}$ is a validation dataset which doesn't contain any attacks.

As the anomaly detection function $\mathcal{A}$ depends on the value of the loss function, we can restate our problem as a bilevel optimization problem:
\begin{equation}
  \label{optimization_problem_outer}
   \mathcal{D}^\star_p \in \argmin_{\mathcal{D}^{'}_p \in \Phi(\mathcal{D}_p)} L(\mathcal{D}_{val} \cup \mathcal{Y}_a, \hat{w})
\end{equation}
\begin{equation}
  \label{optimization_problem_inner}
   s. t.\,  \hat{w} \in \argmin_{w^{'} \in \mathcal{W}} \mathcal{L}(\mathcal{D}_{tr} \cup \mathcal{D}_p, w^{'}),
\end{equation}
where the inner problem is the learning problem, and the function $\Phi$ expresses constraints on the values of the poisoning sequences.
The poisoning sequences influence our objective indirectly, through the parameters $\hat{w}$ of the model trained on them.
The algorithms used for solving this bilevel optimization problem are described in Section \ref{sec:method}.

\section{Related Work}\label{sec:related}
A number of recent studies have focused on evasion attacks on CPS anomaly detectors.
In \cite{feng2017deep}, the authors showed that generative adversarial networks (GANs) can be used for real-time learning of an unknown ICS anomaly detector (more specifically, a classifier) and for the generation of malicious sensor measurements that will go undetected.

The research in \cite{ghafouri2018adversarial} presents an iterative algorithm for generating stealthy attacks on linear regression and feed-forward neural network-based detectors.
The algorithm uses mixed-integer linear programming (MILP) to solve this problem.
For neural network detectors, the algorithm first linearizes the network at each operating point and then solves the MILP problem.
The paper demonstrates a successful evasion attack on a simulated Tennessee Eastman process.

Recently, Erba \etal \cite{erba2019real} demonstrated a successful real-time evasion attack on an autoencoder-based detection mechanism in water distribution systems.
The authors of \cite{erba2019real} considered a white-box attacker that generates two different sets of spoofed sensor values: one is sent to the PLC, and the other is sent to the detector.

The most recent paper in this area \cite{zizzo2019intrusion} also focused on an adversary that can manipulate sensor readings sent to the detector.
The authors showed that such attackers can conceal most of the attacks present in the SWaT dataset.
Our study differs from these studies in a number of ways.
First and foremost, all of the abovementioned papers examined evasion attacks, while our research focuses on poisoning attacks.
Second, \cite{feng2017deep, erba2019real, zizzo2019intrusion} considered a threat model in which the attacker manipulates the detector's input data \textbf{in addition} to manipulating the sensor data fed to the PLC.
Such a model provides a lot of freedom for the adversary to make changes to both types of data.
Our threat model considers a significantly more constrained attacker that can only change the sensory data that is provided \textbf{both} to the PLC and the detector.

A few recently published papers have studied poisoning attacks, however the authors considered them in a different context.
The study performed by Mu\~{n}oz-Gonz\'{a}lez \etal \cite{munoz2017towards} was the  first one to successfully demonstrate poisoning attacks on multiclass classification problems.
It also was the first to suggest generating poisoning data using back-gradient optimization.
Our research extends this method to semi-supervised multivariate time series regression tasks in the online training setting and evaluates the robustness of an autoencoder-based detector to such attacks.

Shafani \etal \cite{shafahi2018poison} and Suciu \etal \cite{suciu2018does} studied clean-label poisoning of classifiers.
In targeted clean-label poisoning, the attacker does not have control of the labels for the training data and changes the classifier's behavior for a specific test instance without degrading its overall performance for other test inputs.
These studies differ significantly from ours, both in terms of the learning task to be poisoned (classification vs. regression) and in the domain (images vs. long interdependent multivariate time sequences).

Madani \etal \cite{madani2018robustness} studied adversarial label contamination of autoencoder-based intrusion detection for network traffic monitoring.
Their research considered a black-box attacker that gradually adds \textbf{existing} malicious samples to the training set, labeling them as normal.
Such a setting is very different from the one studied in our work.
First, we consider semi-supervised training; thus, there is no labeling involved.
Second, we explored algorithms for \textbf{generating} adversarial poisoning samples that will direct the detector's outcome towards the target goal.

In this section, we provided a brief review of related published research.
Although some of the previous research has dealt with related topics or domains, to the best of our knowledge, this study is the first one addressing poisoning attacks on multivariate regression learning algorithms, and specifically on online-trained physics-based anomaly and intrusion detectors in CPSs.

\section{Threat Model}\label{sec:threat_model}
In this study, we consider a malicious sensor threat model widely studied in the wireless sensor network domain (\cite{pires2004malicious, shi2004designing}).
This model was used in the context of adversarial attacks on ICS detectors in \cite{ghafouri2018adversarial} and \cite{kravchik2019efficient} and is illustrated in Figure~\ref{fig:system}.
Under this model, the attacker possesses the knowledge of the historical values measured by the sensors and can spoof arbitrary values of the sensors' readings, however both the PLC and the detector see the same spoofed values.
We selected this model due to its high relevance to the ICS domain.
Consider an ICS with sensors distributed over a large area, which send their data to a PLC residing at a physically protected and monitored location.
In this setup, the adversary can replace the original sensor with a malicious one, reprogram the sensor, change its calibration, influence the sensor externally, or just send false data to the PLC over the cable/wireless connection, but the attacker cannot penetrate the physically protected PLC-to-SCADA network.
We argue that this setup is much more realistic than one in which an attacker controls the internal network of the remote segment or even the network of the control segment considered by \cite{erba2019real} and \cite{zizzo2019intrusion}.
We also argue that our threat model presents more constraints and challenges to the adversary who must achieve his/her goals with a single point of data manipulation.

The attacker's ultimate goal is to carry out a specific attack involving significant changes to the values measured by one or more sensors.
For example, the attacker might aim to report a very low water level in a tank, while in reality the tank is full, thus causing it to overflow.
Without poisoning, the detector would raise an alert upon encountering the spoofed value for the level, as it deviates from the normal system physical behavior patterns learned.
Hence, the goal of the attacker is to poison the detector's model so that after the attack has been launched the detector accepts a spoofed value as normal.

To poison the detector's model, the attacker exploits the detector's \textbf{online learning}.
Unlike in \cite{kravchik2019efficient}, we consider a practice common in ICSs, in which the detector is periodically retrained by using the newly collected data from the monitored system. 
The goal of this online training is to compensate for the concept drift common in physical systems.
We therefore assume that the measured sensor values of the monitored system are added to the training dataset, and thus the attacker can gradually poison the detector to steer it towards the desired attack.
For simplicity, we assume that the attacker knows when to inject the poisoning data so that it will be used for retraining.
We also assume that \textbf{only the data that does not trigger alerts will be used for retraining}.
This is a reasonable assumption, as concept drift is a gradual phenomenon, and the goal of the online training is to allow the detector to adjust to these gradual changes.
This assumption imposes constraints on the level of manipulations the attacker applies to the sensor values.
On the other hand, the attacker can \textbf{add poisoning points gradually}, allowing the system to learn from previous poisoning and thus achieve his/her goal iteratively.

In this study, we assume a white-box attacker that knows the parameters of learning, the learned weights, the detection algorithm, and the hyperparameters, as well as the online training schedule.

\section{Methodology}\label{sec:method}
In this section, we first present the challenges of finding poisoning examples within the context of our research.
We then present two algorithms used to solve this problem and discuss how we applied them to two different modes of data processing: complete signal reconstruction and short-subsequence signal reconstruction.
As mentioned in Section \ref{sec:background_poisoning}, the first challenge is to assess the influence of the poisoning examples on the model's prediction for the attack and validation inputs (Equations~4-5).
The difficulty lies in the fact that the influence is indirect, through the weights updates during the training process, and these updates are not available even for a white-box attacker.
Additional challenges arise from the overlapping signal subsequence modeling common in ICS detectors.
As noted in Section \ref{sec:method_applying_back_gradient} where this challenge is discussed further, this will force the attacker to consider the very long-term impact of the poisoning on the manipulated signal.
To clarify the terminology, in this section and those that follow, we refer to a single manipulated sequence of feature vectors as a \textbf{poisoning point}.
As we are looking for a \textbf{sequence} of \textit{sequences} that will achieve the desired adversarial goal, calling the individual sequences \textbf{poisoning points} helps distinguish between the outcome of a single iteration of the poisoning algorithm (\textbf{a poisoning point)} and the resulting set of poisoning points.

\subsection{Finding a poisoning sequence with back-gradient optimization}\label{sec:method_back_gradient}
First, we note that in order to reach the attacker's goal (Equation~\ref{objective_func}), the attacker would proceed by starting with an empty set and iteratively finding the next poisoning point by solving the bilevel optimization problem.
Thus, each iteration step can be rewritten as solving
\begin{equation}
  \label{optimization_problem_outer_simple}
   y^\star_c \in \argmin_{y^{'}_c \in \Phi(y_c)} \mathcal{A} (y^{'}_c) = L(\mathcal{D}_{val} \cup \mathcal{Y}_a, \hat{w})
\end{equation}
\begin{equation}
  \label{optimization_problem_inner_simple}
   s. t.\,  \hat{w} \in \argmin_{w^{'} \in \mathcal{W}} \mathcal{L}(\mathcal{D}_{tr} \cup y^{'}_c, w^{'}),
\end{equation}
or, in a simplified form,
\begin{equation}
  \label{optimization_problem_outer_simplest}
   y^\star_c \in \argmin_{y^{'}_c \in \Phi(y_c)} \mathcal{A} (y^{'}_c) = L(\mathcal{Y}_a, \hat{w})
\end{equation}
\begin{equation}
  \label{optimization_problem_inner_simplest}
   s. t.\,  \hat{w} \in \argmin_{w^{'} \in \mathcal{W}} \mathcal{L}(y^{'}_c, w^{'}),
\end{equation}
where $y_c$ is the poisoning point optimized in the current iteration.
The iterations stop when Equation~ \ref{objective_func} is satisfied. 

One approach for solving Equations~\ref{optimization_problem_outer_simplest}-\ref{optimization_problem_inner_simplest} is to use gradient ascent \cite{munoz2017towards}:
\begin{equation}
  \label{eq:gradient_ascent}
    \nabla_{y_c}\mathcal{A} = \nabla_{y_c}L + {\frac{\delta \hat{w}}{\delta y_c}}^T \nabla_{w}L.
\end{equation}

As neither the validation set $\mathcal{D}_{val}$ nor the attack input contains the poisoning points, there is no explicit dependency between the attacker's objective $\mathcal{A}$ and the poisoning points.
Therefore, the first part of Equation~\ref{eq:gradient_ascent} is equal to zero, and our goal is to find the value for $\frac{\delta \hat{w}}{\delta y_c}$ that reflects the influence of the poisoning point on the attacker's objective through the learned weights.
However, it is impossible to use direct numeric methods to calculate this influence, as the weights are updated multiple times during the learning process, and their intermediate values are not stored and are thus not available to the adversary.
To cope with this problem, we use back-gradient optimization \cite{maclaurin2015gradient}, as suggested in \cite{munoz2017towards}.
The core idea of this approach is iterative backwards calculation of both the weights' updates and  $\frac{\delta \hat{w}}{\delta y_c}$, performed by reversing the learning process and calculating the second gradients in each iteration. 

In our research, we implemented back-gradient optimization for stochastic gradient descent according to Algorithm \ref{alg_back_gradient_descent} (based on \cite{munoz2017towards}).
\begin{algorithm}[t!]
\caption{Find $\nabla_{y_c}\mathcal{A}$ given trained parameters $w_T$, learning rate $\alpha$, attack input $y_a$, poisoning point $y_c$, loss function $L$, and learner's objective $\mathcal{L}$,  using back-gradient descent for T iterations.}
\label{alg_back_gradient_descent}
\begin{algorithmic}[1]
\Function{getPoisonGrad}{$w_T, \alpha, y_a, y_c, L, \mathcal{L}$}
\State $dy_c \gets 0$
\State $dw \gets \nabla_{w}L(y_a, w_T)$
\For {$t = T$ to 1} 
\State $dy_c \gets dy_c - \alpha dw  \nabla_{y_c}  \nabla_{w} \mathcal{L}(y_c, w_t)$
\State $dw \gets dw - \alpha dw  \nabla_{w}  \nabla_{w} \mathcal{L}(y_c, w_t)$
\State $g_{t-1} \gets \nabla_{w} \mathcal{L}(y_c, w_t)$
\State $w_{t-1} \gets w_t +  \alpha g_{t-1}$
\EndFor
\State \textbf{return} $dy_c$
\EndFunction
\end{algorithmic}
\end{algorithm}
Algorithm~\ref{alg_back_gradient_descent} starts with initializing the derivatives of the loss relative to the attack input and the weights of the trained model (lines 2 and 3).
Then it iterates for a given number $T$ iterations, rolling back the weights' updates made by the training optimizer (lines 7 and 8).
In each iteration, the algorithm calculates the second derivatives of the loss relative to the weights and the attack input at the current weights' values (lines 5 and 6) and updates the  values maintained for both derivatives.
The final value of $\nabla_{y_c}\mathcal{A}$ accumulates the compound influence of the poison input through the weights' updates. 
Calculating the second derivatives is very expensive computationally, therefore Hessian-vector products were used to optimize the calculation of $dw \nabla_{y_c}  \nabla_{w} \mathcal{L}$ and $dw \nabla_{w} \nabla_{w} \mathcal{L}$, as proposed in \cite{pearlmutter1994fast}.

\subsection{Applying back-gradient optimization to periodic signals}\label{sec:method_applying_back_gradient}

Algorithm \ref{alg_poisoning}, which is one of the contributions of this research, was used to apply Algorithm~\ref{alg_back_gradient_descent} to autoregression learning of periodic signals.

Algorithm \ref{alg_poisoning} starts with an empty set of poisoning points (line 1) and an initial poisoning value.
Then it repeatedly uses a $train\_test$ function to perform the model retraining with the current training and poisoning datasets (line 6).
If the target attack input and the clean validation data do not raise alerts, the problem is solved (lines 7-9).
Otherwise, the gradient of the poisoning input is calculated using Algorithm~\ref{alg_back_gradient_descent}, normalized, and used to find the next poison value (lines 10-11).
The new poison value is tested with the current detector (line 13).
If it raises alerts, the value is too large, and the last poison value that did not raise an alert is added to $\mathcal{D}_p$, the adversarial learning rate is decreased, and the last good (capable of being added without raising an alert) poison is used as a base for the calculation in the next iteration (lines 14-17).
If the learning rate becomes too low, the algorithm terminates prematurely (lines 18-19).
If no alerts were raised, the learning rate is restored to its original value for the next iteration, in order to accelerate the poisoning progress (lines 20-21). 
\begin{algorithm}[t!]
\caption{Find poisoning sequence set $\mathcal{D}_p$ given $\mathcal{D}_{tr}, \mathcal{D}_{val}$, learning rate $\alpha$, adversarial learning rate $\lambda$, attack input $y_a$, initial poisoning point $y^0_c$, loss function $L$, learner's objective $\mathcal{L}$, and maximum number of iterations M.}
\label{alg_poisoning}
\begin{algorithmic}[1]
\State $\mathcal{D}_p \gets []$
\State $decay \gets 0.9$
\State $eps \gets 0.00001$ \Comment {Minimal allowed $\lambda$}
\State $orig\lambda \gets \lambda $
\For {$i = 1$ to M} 
\State $(w_T, alerts) \gets train\_test(\mathcal{D}_{tr}, \mathcal{D}_{val}, y_a, \mathcal{D}_p, y^i_c) $
\If{$alerts == 0$}
\State ${D}_p \pluseq (y^{i}_c)$ \Comment {Add current poison}
\State \textbf{break}
\EndIf
\State $dyc \gets getPoisonGrad(w_T, \alpha, y_a, y^i_c, L, \mathcal{L})$
\State $y^{i+i}_c \gets y^{i}_c - \lambda \cdot dyc / \max(dyc)$
\State\Comment{Check that the new poison does not generate alerts}
\State $(w_T, alerts) \gets train\_test(\mathcal{D}_{tr}, \mathcal{D}_{val}, y_a, \mathcal{D}_p, y^{i+i}_c) $
\If{$alerts > 0$}
\State ${D}_p \pluseq (y^{i}_c)$ \Comment {Add previous poison}
\State $\lambda \gets decay\cdot\lambda$ \Comment {Adjust learning rate}
\State $y^{i+i}_c \gets y^{i}_c$ \Comment {Revert to last good poison}
\If{$\lambda <= eps$}
\State \textbf{break}\Comment {Can't find anymore poisons}
\EndIf
\Else
\State $\lambda \gets orig\lambda$ 
\EndIf
\EndFor
\State \textbf{return} $\mathcal{D}_p$
\end{algorithmic}
\end{algorithm}

For simplicity, we omitted the adversarial learning rate's ($\lambda$) dynamic decay used in implementation from the description of Algorithm~\ref{alg_poisoning}.
With the dynamic decay, $\lambda$ is decreased if the test error has not decreased and the iterations are terminated early if $\lambda < 0.00001$.
Another implementation optimization not shown in the pseudocode of Algorithm~\ref{alg_poisoning} adds \textit{clean} data sequences to $\mathcal{D}_p$ if the detector raises alerts on validation data.
If this happens, the model is ``over-poisoned'' and the clean data is added until the validation alerts disappear. 
This is done after the calls to $train\_test$.

Once the validity of Algorithm \ref{alg_poisoning} was confirmed, we needed to adapt it to the way neural networks are applied to continuous signals in anomaly and attack detectors (e.g., \cite{lin2018tabor}, \cite{kravchik2018detecting}, \cite{taormina2018deep}, \cite{kravchik2019efficient}, and many others).
The common practice is to apply the neural networks to the multivariate sequences formed by sliding a window of a specified length over the input signal.
These sequences are overlapping, hence a single time point appears in multiple sequences.
As a result, a change to a single time point by an attacker affects the detector's predictions for the multiple sequences that include this point.
Moreover, in order for the changed point to remain undetected, its prediction should also be close to its (changed) value based on multiple past input sequences.
These self-dependencies spread across time, both forward and backwards, and must be taken into account when creating the poisoning input, as this input must therefore be much longer than the sequence of points changed during the target attack .
However, the model at the attacker's disposal deals only with the short sequences.
In order to be able to evaluate the total loss value of the attack for the entire input, we performed the optimization on a \textbf{wrapper model ($\mathcal{WM}$}) built around the original trained model.
The wrapper model allows for calculating the gradients and optimizing the adversarial input for an arbitrary long input sequence, similar to the process described in \cite{kravchik2019efficient}.
\begin{figure}[t]
\centering{\includegraphics[clip, trim=0cm 0cm 0cm 0cm, scale=0.35]{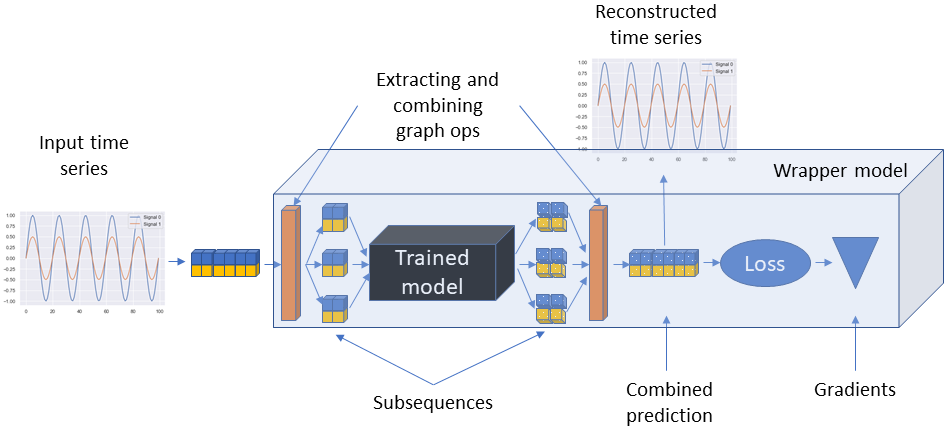}}
\caption{Wrapper model.}
\label{fig:wrapper_model}
\end{figure}
The wrapper model illustrated in Figure \ref{fig:wrapper_model} extends the trained model's graph to calculate the gradient of the attacker's objective relative to the entire input.
Specifically, the wrapper model prefixes the original model with graph operations that divide the long input into overlapping subsequences and appends the model with operations that combine the results of individual predictions and calculate the combined output.
The wrapper model allows us to use  Algorithm \ref{alg_poisoning} and Algorithm \ref{alg_naive_poisoning} (described in Section~\ref{sec:method_baseline_naive_alg}), as is, to poison multivariate time series data in a setting in which ICS anomaly detectors are commonly deployed.

\subsection{Initial poison choice}\label{sec:method_initial_poison_choice}
Proper choice of the initial poisoning point is an important factor in the performance of Algorithm \ref{alg_poisoning}.
As the loss surface of the optimized function is not smooth, some initialization points will cause the algorithm to get stuck in local minimum solutions.
We explored two approaches for the initial poison choice: a benign-data-based approach and an attack-based approach.
The benign-data-based approach uses the original unpoisoned data as the initial poisoning value.
For the synthetic signals, it uses the pure generated signal (e.g., sine wave).
For the real data, it initializes the poison with the values of the sensors when not under attack.

The attack-based approach uses the sensors' values that are closest to the sensors' values under attack but do not cause an alert for the initial poison. 
This approach is more aggressive and allows the areas with the local minimum of the loss function, which the good-data-based approach is prone to get stuck in, to be skipped.
The attack-based approach was used in the real-world data experiment we conducted in this research (as described below in Section \ref{subsec:results_swat}).
Without physical access to the SWaT testbed, we used the SWaT public dataset to create the poisoning inputs for the attacks it contains.
In this setting, we had no access to the ``untainted'' data values that would appear if the attack had not happened.
The attack-based approach proved to work much better than other techniques tested, such as finding the most similar sequence in the training dataset and using the detector's trained model to generate the initial poison (similar to the approach of \cite{erba2019real})
In order to determine the initial poisoning based on the desired attack to conceal, we utilized \textbf{gradient ascent}, starting from the target attack sequence and iteratively moving the poisoning point in the direction of $-\nabla_{y_c}L$.
As above, the optimization was applied on the wrapper model.

\subsection{Interpolative poisoning algorithm}\label{sec:method_baseline_naive_alg}
In addition to the back-gradient optimization algorithm (Algorithm \ref{alg_back_gradient_descent}), we propose a much simpler naive interpolation algorithm to find the poisoning sequence.

This algorithm is based on an observation that both the initial poisoning point and the final attack point are known in advance.
Similarly to Algorithm \ref{alg_back_gradient_descent}, the interpolative Algorithm \ref{alg_naive_poisoning} starts with an empty set of poisoning points, an initial poisoning point, and an initial interpolation step (lines 1-5).
In each iteration, the algorithm attempts to add a poisoning point that is an interpolation between the initial point and the final point (lines 7-9).
If the new poisoning point does not raise an alert, it is added to the result set, and the next interpolation between it and the target attack is tested (lines 12-15).
Otherwise, the interpolation step is decreased and the interpolation is recalculated (lines 10-11).
The algorithm continues until success is achieved or the interpolation step becomes too small.
As in Section \ref{sec:method_applying_back_gradient}, the algorithm's implementation included the addition of clean data points to $\mathcal{D}_p$ if the model is over-poisoned and causes alerts on validation data.
This optimization is omitted from the pseudocode for brevity.

This concludes our description of our algorithms and methodology, and the next section presents the experimental results obtained by applying them to both synthetic and real-world data.

\begin{algorithm}[t!]
\caption{Find poisoning sequence set $\mathcal{D}_p$ given $\mathcal{D}_{tr}, \mathcal{D}_{val}$, decay rate $\delta$, attack input $y_a$, and initial poisoning point $y^0_c$.}
\label{alg_naive_poisoning}
\begin{algorithmic}[1]
\State $eps \gets 0.0000001$
\State $\mathcal{D}_p \gets []$
\State $rate \gets 1$
\State $step \gets 1$
\State $y_p \gets y^0_c$
\While {$\max(\left|step\right|) > eps$} 
\State $step = rate \cdot (y_a - y_p)/2$
\State $y_c = y_p + step$
\State $(err, alerts) \gets train\_test(\mathcal{D}_{tr}, \mathcal{D}_{val}, y_c, \mathcal{D}_p) $\Comment {Test if current poison raises alert}
\If{$alerts$}
\State $rate \gets rate \cdot \delta$ \Comment {Decrease the rate}
\Else
\State $y_p \gets y_c $ \Comment {Start interpolating from the new point}
\State ${D}_p \pluseq y_c$ \Comment {Add current poison}
\State $rate \gets rate/\delta$ \Comment {Increase the rate}
\State $(err, alerts) \gets train\_test(\mathcal{D}_{tr}, \mathcal{D}_{val}, y_a, \mathcal{D}_p) $\Comment {Test if the attack raises alert after poisoning}
\EndIf
\If{$alerts == 0$}
\State \textbf{break}\Comment {The goal is reached}
\EndIf
\EndWhile
\State \textbf{return} $\mathcal{D}_p$
\end{algorithmic}
\end{algorithm}

\section{Experiments and Results}\label{sec:results}
In this section, we first evaluate the effectiveness of the interpolation and back-gradient algorithms in executing poisoning attacks on synthetic periodic signals.
Then we explore the influence of different factors on the ability to poison the given model, and finally, we present the results of producing poisoning samples for attacks from the popular SWaT dataset.

\subsection{Detector, training procedure and evaluation criteria}\label{sec:results_model_training}
\begin{figure}[t]
\centering{\includegraphics[clip, trim=0cm 0cm 0cm 0cm, scale=0.36]{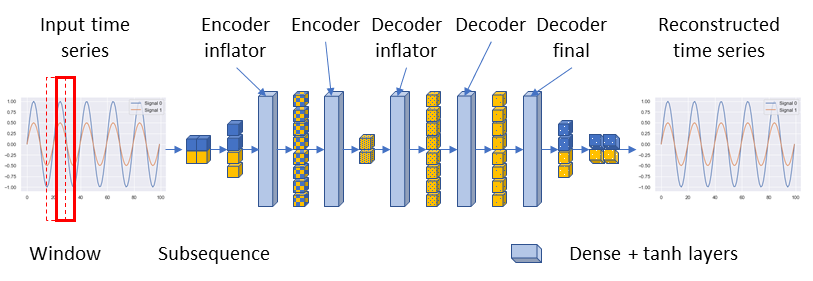}}
\caption{Autoencoder architecture used in ICS attack detector.}
\label{fig:model}
\end{figure}

A simple undercomplete autoencoder (UAE) network was used for the ICS detector under test.
We used the network architecture described in \cite{kravchik2019efficient} for all tests, with both synthetic and real data.
The simplest instance of such a detector model is presented in Figure~\ref{fig:model}.

The single difference of this architecture from the classic UAE is the presence of inflating layers before both the encoder and decoder parts.
As explained in \cite{kravchik2019efficient}, these layers improve the model performance by increasing its hypothesis space. 

The detector was implemented in TensorFlow and trained using the gradient descent optimizer until the test error (measured as the mean squared reconstruction error for the input signal) decreased to less than 0.01.
While we experimented with various numbers of encoder and decoder layers, and multiple inflation factors and input-to-code ratios, these variables mainly influenced the detector's accuracy and not the poisoning results.
The results below are for the tests performed with an inflation factor of two, an input-to-code ratio of two, a single encoding and decoding layer, and a subsequence length of two.

Our study focuses on online-trained ICS detectors.
In the online training mode, the model is periodically retrained with part of the previously used training data blended with the newly arrived data.
For simplicity, we performed model retraining with newly generated poisoning input after each poisoning iteration, with the new poisoning input appended to the existing training data.
There are other possible ways of combining the new and existing training data, e.g., randomly selecting a fixed number of data samples from both.
We experimented with this setup as well and discovered that it causes a larger number of poisoning points to be added but does not change the overall findings.

The following metrics were used for the poisoning effectiveness evaluation:
\begin{itemize}
    \item the \textbf{attack magnitude}, measured as the maximal difference from the original and the target spoofed sensor value;
    \item the \textbf{number of poisoning points} in the generated sequence;
    \item the \textbf{number of optimization iterations} required to find the poisoning sequence.
\end{itemize}

We ran grid search tests for both poisoning algorithms, with and without the initial poison optimization described in Section~\ref{sec:method_initial_poison_choice} for multiple values of:
\begin{itemize}
\item the training set size, 
\item target attack magnitude, 
\item attack location, 
\item input signal length,
\item subsequence length,
\item training and adversarial iterations.
\end{itemize}

\subsection{Data}\label{subsec:results_data}

\begin{figure}[t]
\centering{\includegraphics[clip, trim=0cm 0cm 0cm 0cm, scale=0.36]{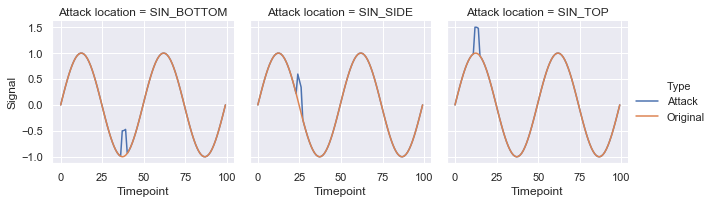}}
\caption{Different kinds of attack locations relative to the signal's period.}
\label{fig:attack_locations}
\end{figure}

\textbf{Synthetic dataset.}
For the synthetic data experiments we used a number of simple periodic signals (sine, cosine, square function (numpy.square), and saw tooth function (numpy.sawtooth)) of different time periods and lengths.
The signal amplitude was between $-1$ and 1, and a distorting Gaussian noise with a mean of 0 and a standard deviation of 0.05 was applied to the signal.
To simulate the attacks, we increased the signal amplitude by a specified value.
A number of different attack locations were tested, such as the highest point of the signal (SIN\_TOP), the lowest point of the signal (SIN\_BOTTOM), and the middle of the slope (SIN\_SIDE), as illustrated in Figure~\ref{fig:attack_locations}.
The rational behind testing various attack locations was to model various adversaries' attack objectives.

For multiple synthetic signals, we used a number of linear dependent sines to model a simple case of system characteristics correlated to each other due to the laws of physics, such as water flow and water level.

\textbf{ICS testbed dataset. }
For the real-world data experiments, we utilized the popular SWaT dataset~\cite{goh2016dataset}.
The dataset was collected from the Secure Water Treatment (SWaT) testbed at the Singapore University of Technology and Design and has been used in many studies since it was created.
The testbed is a scaled-down water treatment plant, running a six-stage water purification process.
Each process stage is controlled by a PLC with sensors and actuators connected to it.
The sensors include flow meters, water level meters, and conductivity analyzers, while
the actuators are water pumps, chemical dosing pumps, and inflow valves.
The dataset contains 51 attributes capturing the states of the sensors and actuators for every second for seven days of recording under normal conditions and four days of recording when the system was under attack (the data for this time period contains 36 attacks).
Each attack targets a concrete physical effect, such as overflowing a water tank by falsely reporting a low water level, thus causing the inflow to continue.

Following our threat model (Section \ref{sec:threat_model}), we selected the attacks that involved sensor value manipulations.
Table \ref{tab:swat_poisoning_att} lists and describes the SWaT attacks chosen for the poisoning experiments.

\begin{table}[t]
\begin{center}
\begin{threeparttable}
\caption{SWaT attacks selected for poisoning.}
\label{tab:swat_poisoning_att}
\begin{tabular}{|c|m{3.9em}|m{6.1em}|m{6em}|m{7.26em}|}
 \hline
  \# & Attacked sensor(s) & Start state & Description & Expected impact\\
 \hline
 3 & LIT-101 & Water level between L and H & Increase by 1mm every second & Tank underflow; Damage P-101\\
 \hline
 7 & LIT-301 & Water level between L and H & Water level increased above HH & Stop of inflow; Tank underflow; Damage P-301\\
 \hline
 8 & DPIT-301 & Value of DPIT is $<40$kpa & Set value of DPIT as $>40$kpa & Backwash process is repeatedly started; Normal operation stops\\
 \hline
 10 & FIT-401 & Value of FIT-401 above 1 & Set value of FIT-401 as $<0.7$ & No	UV shutdown; P-501 turns off\\
 \hline
 11	& FIT-401 &	Value of FIT-401 above 1 & Set value of FIT-401 as 0 &	UV shutdown; P-501 turns off\\
\hline
16	& LIT-301 &	Water level between L and H &	Decrease water level by 1mm each second & Tank overflow\\
\hline
31 & LIT-401 &	Water level between L and H &	Set LIT-401 to less than L & Tank overflow \\
\hline
32 & LIT-301 &	Water level between L and H &	Set LIT-301 to above HH	& Tank underflow; Damage P-302\\
\hline
33 & LIT-101 &	Water level between L and H &	Set LIT-101 to above H	& Tank underflow; Damage P-101\\
\hline
36 & LIT-101 &	Water level between L and H &	Set LIT-101 to less than LL & Tank overflow \\
\hline
\end{tabular}
\begin{tablenotes}
\small
\item Legend: L - low setpoint value, H - high setpoint value, LL - dangerously low level, HH - dangerously high level. 
\end{tablenotes}
\end{threeparttable}
\end{center}
\vspace{-6mm}
\end{table}

There were five other sensor-spoofing attacks in the SWaT dataset, however they involved features that did not preserve their statistics between the training and the test datasets and thus could only be modeled very poorly, as shown in \cite{kravchik2019efficient}, and were therefore not included in our experiment.
\begin{figure}[t]
\centering{\includegraphics[clip, trim=0cm 0cm 0cm 0cm, scale=0.62]{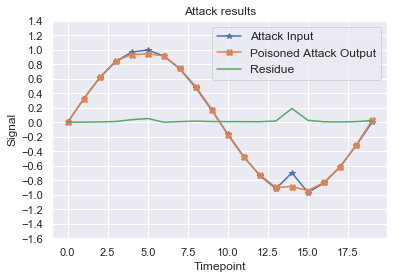}}
\caption{Successful poisoning results. 
The detector predicts a signal that is close enough to the attack signal, and the residue is below the threshold of 0.2.}
\label{fig:poisoned_attack}
\end{figure}
\subsection{Synthetic signal poisoning}\label{subsec:results_synthetic_single}
The goals of the experiments on synthetic signals were to assess the algorithms' effectiveness and to examine the influence of different test parameters on the algorithms' poisoning ability.
The results of our experiments conducted with two kinds of signals (a sine and a double sine) are presented below.
All of the experiments were conducted with the anomaly detection threshold of 0.2, using the SIN\_BOTTOM attack location (unless stated otherwise), with the model and test setup described in Section \ref{sec:results_model_training}.

Figure~\ref{fig:poisoned_attack} illustrates the results of successful poisoning - given an input signal of a sine wave distorted by the attack in the SIN\_BOTTOM location, the poisoned model produces a prediction whose deviation from the attack signal is less than the threshold of 0.2.

\begin{figure}[t!]
\centering{\includegraphics[clip, trim=0cm 0cm 0cm 0cm, scale=0.62]{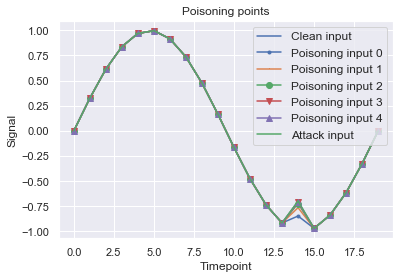}}
\caption{Poisoning points generated by the interpolative algorithm.}
\label{fig:poisoning_points}
\end{figure}
Figure~\ref{fig:poisoning_points} demonstrates the sequence of the poisoning points leading to successful poisoning produced by the interpolative algorithm.
One can observe that the poisoning points form a successive interpolation between the original unpoisoned signal and the target attack signal.

\begin{figure*}[!htb]
\centering{\includegraphics[clip, trim=0cm 0cm 0cm 0cm, scale=0.6]{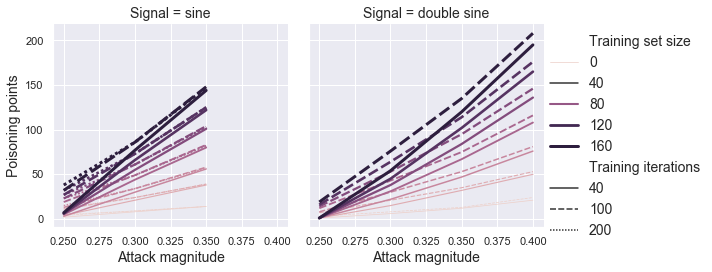}}
\caption{Dependency of the number of required poisoning points on the attack magnitude for single-sequence synthetic data with the interpolative algorithm. The number of points increases nearly linearly with the attack amplitude and depends strongly on the training set size.}
\label{fig:single_amplitude}
\end{figure*}

The training and test datasets consisted of period-aligned sine signals distorted by Gaussian noise (as described in Section \ref{subsec:results_data}).

We started by studying poisoning for relatively long (100 timepoints) independent signals that were modeled at once, without breaking them into short overlapping subsequences (subsequently called \textit{single-sequence data}).
The experiments showed that the \textbf{interpolative} algorithm was able to successfully poison a trained model for attacks up to the amplitude of 0.4, and the amount of required poisoning samples increased linearly with the target attack amplitude, as shown in Figure~\ref{fig:single_amplitude}.
We found that \textbf{more poisoning samples than training samples} were required  in order to reach the amplitude of 0.4, which is an extremely high ratio.
It was impossible to poison the model for larger attack amplitudes without triggering an alert either on the poisoning input or on a clean validation sample.
As Figure~\ref{fig:single_amplitude} shows, the number of training iterations did not have a strong influence on the poisoning results.

\begin{figure}[htpb]
\centering{\includegraphics[clip, trim=0cm 0cm 0cm 0cm, scale=0.4]{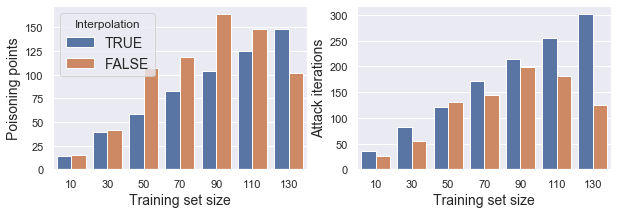}}
\caption{Comparison of the number of poisoning points and attack iterations for the interpolative and back-optimization algorithms for single-sequence data.}
\label{fig:naive_vs_backop_single}
\end{figure}

For this single-sequence modeling, the \textbf{back-gradient optimization} algorithm reached the same attack magnitudes, but in some cases required more poisoning points than the interpolative one, as demonstrated in Figure~\ref{fig:naive_vs_backop_single}.

\begin{figure}[htpb]
\centering{\includegraphics[clip, trim=0cm 0cm 0cm 0cm, scale=0.45]{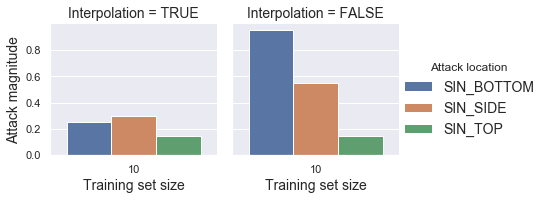}}
\caption{Comparison of maximal attack magnitude reached by poisoning for different attack locations. The attack was performed on multi-sequence data.}
\label{fig:max_poison_locations}
\end{figure}

\begin{figure}[htpb]
\centering{\includegraphics[clip, trim=0cm 0cm 0cm 0cm, scale=0.6]{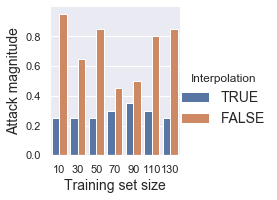}}
\caption{Comparison of maximal attack magnitude reached by the interpolative and back-optimization algorithms poisoning for multi-sequence data and SIN\_BOTTOM attacks.}
\label{fig:max_poison_multi}
\end{figure}

After validating the ability of both algorithms to achieve the target poisoning in the single-sequence setting, we tested their performance with detectors, based on modeling multiple overlapping short subsequences as commonly done in ICS attack detectors, as described in Section~\ref{sec:method_applying_back_gradient}.
We refer to this setup as \textit{multi-sequence data} in the text below.
For multi-sequence data, we observed that the ability to poison the model strongly depends on the attack location.
As shown in Figure~\ref{fig:max_poison_locations}, neither algorithm was able to generate poisoning input for any significant attack in the SIN\_TOP location.
The attacks in the SIN\_TOP location aim to spoof the sensor value, increasing it beyond its highest value and thus trigger the response of the PLC; such attacks are usually the most valuable to the attacker.
As evident from Figure~\ref{fig:max_poison_locations}, for the SIN\_BOTTOM and SIN\_SIDE locations, the back-optimization algorithm was able to reach significantly greater attack magnitudes than the interpolative one.
This difference was consistent across different training set sizes, as can be see in Figure~\ref{fig:max_poison_multi}.
The reason for this phenomenon is the ability of back-gradient optimization to find local minima of the loss function which are missed by the relatively large steps taken by the interpolative algorithm.

\begin{figure}[htpb]
\centering{\includegraphics[clip, trim=0cm 0cm 0cm 0cm, scale=0.4]{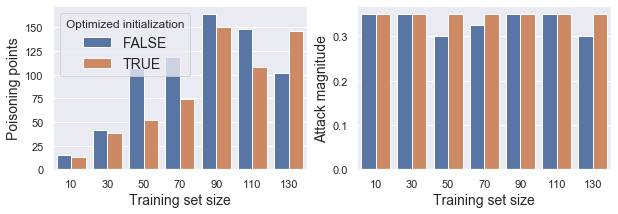}}
\caption{The influence of initial poison sequence optimization. The attack was performed in the SIN\_BOTTOM location on multi-sequence data.}
\label{fig:find_poison_effect}
\end{figure}

We also tested the influence of the gradient ascent poison initialization described in Section \ref{sec:method_initial_poison_choice} on the performance of the back-gradient algorithm.
Figure~\ref{fig:find_poison_effect} presents the impact of this initialization on the amount of poisoning points required to reach the maximal attack magnitude and the size of the maximal attack.
It can be seen that the optimized initialization allowed us to achieve the target poisoning with less poisoning points, while maintaining the target attack magnitude, and even increasing it in some cases.

Finally, we  note that the execution time of the interpolative algorithm's iteration is significantly shorter than the execution time of the back-gradient algorithm, as shown in Table~\ref{tab:timing}. 
\begin{table}[t]
\begin{center}
\begin{threeparttable}
\caption{Iteration execution time (in seconds).}
\label{tab:timing}
\begin{tabular}{|c|c|c|}
 \hline
   Model & Interpolative & Back-gradient\\
 \hline
   Single sequence & 1.27 & 33.16\\
 \hline
   Multiple sequences & 1.24 & 67.37\\
 \hline
\end{tabular}
\begin{tablenotes}
\small
\item The tests were run on Google's Colab TPU for sequences of 100 timepoints. 
\end{tablenotes}
\end{threeparttable}
\end{center}
\vspace{-6mm}
\end{table}

To summarize, the tests conducted on the synthetic data demonstrated that both of the proposed algorithms can poison autoencoder models.
The interpolative algorithm is much faster, but the back-gradient optimization algorithm produces superior results in the learning setup usually deployed for physics-based anomaly detection in ICSs.
At the same time, both algorithms required a long time and a very large number of poisoning points to produce the desired poisoning unless the target attack was of a small magnitude and the training data set was small.
In addition, both algorithms failed to produce poisoning for even low magnitude attacks for the SIN\_TOP attack location. 

\subsection{Poisoning Attacks on SWaT}\label{subsec:results_swat}
After validating the effectiveness of the algorithms, we applied them to the ten relevant attacks from the SWaT dataset described in Section~\ref{subsec:results_data}.
We started by modeling the attacked sensor's signal, along with an additional related signal.
For example, attack \#3 (see Table~\ref{tab:swat_poisoning_att}) spoofs the water level sensor LIT-101 which is installed in the first water tank. 
We modeled sensor LIT-101 along with FIT-101, which measures the water flow into the same tank.
The NN model used had the autoencoder architecture described in Section \ref{sec:results_model_training}.

As the SWaT database contains almost one million records, there was a need to choose the amount of training data to use for model retraining.
In online training, the model is retrained with the most recent data, therefore we used the last part of the training data and appended the generated poisoning points to it.
Considering the linear dependency of the required poisoning points on the training set size, we selected a training set length that was ten times longer that the desired attack sequence, striking a balance between the test's representativeness and its running time.
We subsampled the SWaT dataset at a rate of five seconds and normalized both the training and test data into the 0-1 interval using the training data as a normalization base.

\begin{figure}[htpb]
\centering{\includegraphics[clip, trim=0cm 0cm 0cm 0cm, scale=0.4]{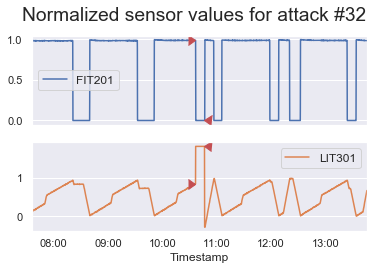}}
\caption{Attack \#32. The red arrows indicate the attack duration. During the attack, the normalized value of the spoofed sensor reached 1.814.}
\label{fig:attack_32}
\end{figure}

As evident from the synthetic test results, the attack location heavily influences poisoning.
The analysis of the selected SWaT attacks revealed that most of the attacks belong to the SIN\_TOP class, as in these attacks the spoofed value is set far beyond the maximum value of the same sensor in the training dataset; this is illustrated in Figure~\ref{fig:attack_32}.
The value of the LIT301 sensor is set to 1.814 during the attack, while 1.0 is the highest value seen during training.

In addition, most of the attacks have a large magnitude (which we approximated by the deviation from the normalization limits, namely below zero or above one), as shown in Table~\ref{tab:swat_poisoning_res}.
As demonstrated in Section~\ref{subsec:results_synthetic_single}, high magnitude values make poisoning impossible, therefore we clipped the attacks to lower values in order to evaluate the feasibility of poisoning with more moderate attacks.

\begin{table}[t]
\begin{center}
\begin{threeparttable}
\caption{SWaT attacks' poisoning results.}
\label{tab:swat_poisoning_res}
\begin{tabular}{|c|m{4em}|m{4em}|m{4em}|m{4em}|m{4em}|}
 \hline
  \# & Modeled sensor(s) & Full magnitude & Clipped magnitude & Poisoned & Poisoning points\\
 \hline
 3 & LIT-101, FIT-101 & 1.325 & 0.25 & No & -\\
 \hline
 7 & LIT-301, FIT-201 & 1.809 & 0.25 & No & -\\
 \hline
 8 & DPIT-301, P-602 & 2.19 & 0.25 & No & -\\
 \hline
 10-11 & FIT-401, LIT-401 & 22.597 & 0.25 & No & -\\
\hline
 16	& LIT-301, FIT-201 & 0.893 & 0.55\tnote{1} & Yes & 2\\
\hline
16 & LIT-301, FIT-201, FIT-301, P-302 & 0.893 & 0.25 & No & -\\ 
\hline
31 & LIT-401, FIT-401 & 0.769 & 0.35 & Yes & 0\tnote{2}\\
\hline
31 & LIT-401, FIT-401 & 0.769 & 0.25 & No & -\tnote{3}\\
\hline
32 & LIT-301, FIT-201 & 1.814 & 0.25 & No & -\\
\hline
33 & LIT-101, FIT-101 &	0.582 & 0.4\tnote{1} & Yes & 1\\
\hline
33 & LIT-101, FIT-101, MV-101, P-101 &	0.582 & 0.25 & No & -\\
\hline
36 & LIT-101, FIT-101 &	1.213 & 0.4\tnote{1} & Yes & 1\\
\hline
36 & LIT-101, FIT-101, MV-101, P-101 &	1.213 & 0.25 & No & -\\
\hline
\end{tabular}
\begin{tablenotes}
\small
\item The maximal attack magnitude achieved by poisoning is presented. A threshold of 0.2 was used in the experiment. The absolute values of magnitudes are presented. The clipping magnitudes increased with steps of 0.05.
\item[1]Achieved by back-gradient optimization.
\item[2]Residue was less than the threshold of 0.2. 
\item[3]Performed with the threshold of 0.1. 
\end{tablenotes}
\end{threeparttable}
\end{center}
\vspace{-6mm}
\end{table}

The test results are presented in Table~\ref{tab:swat_poisoning_res}.
The results demonstrate both the validity of the proposed algorithms and the robustness of autoencoders to poisoning.
The ability of the algorithms to find poisoning input for real-world data serves as a  proof of the algorithms' validity.
In only four of the ten attacks the algorithms were able to find poisoning input that allowed the attack  to be carried out without triggering an alert or introducing false positives for valid data.
In all of the successful cases, the back-gradient algorithm achieved greater attack magnitudes than the interpolative one.
At the same time, it is evident that the successful attacks' magnitude was low in absolute value and substantially lower than the original attacks present in the dataset.

On the other hand, for the majority of the attacks, no poisoning could be found.
Furthermore, all four of the initially successful poisonings were easily eliminated in successive tests.
For three attacks, namely 16, 33, and 36, adding just two additional related sensors to the model prevented poisoning completely.
The majority of the ICS detectors proposed in previous publications
\cite{lin2018tabor, kravchik2018detecting, taormina2018deep, kravchik2019efficient} modeled multiple fields of the monitored system and in some cases combined  all of the fields.
Consequently, we can conclude that adding more fields is the recommended way of using autoencoders in ICS IDSs and that doing so further increases their adversarial poisoning robustness.
For the remaining successfully poisoned attack (\#31), the residue of the model's prediction for the attack was lower than a selected threshold of 0.2, therefore the poisoning process had not started.
When we lowered the threshold to 0.1, neither poisoning algorithm could produce the desired poisoning.

Overall, the experiments on the attacks from the SWaT dataset demonstrate the strong resilience of the tested NN architecture to adversarial poisoning.
The experiments indicate that unlike other problem domains (e.g., image classification), physics-based NN-based anomaly and attack detectors \textbf{cannot easily be  manipulated}.
Even a long optimization procedure cannot cause such detectors to accept a significant anomaly as normal without generating false positives on the normal data.

\section{Discussion and Conclusions}\label{sec:conclusions}
ML-based autonomous systems promise enormous benefits to society.
One of the major obstacles to their increased adoption is the inability to fully trust ML-based decisions, due, to a large extent, to the vulnerabilities of such systems to adversarial attacks.
In the ICS IDS context, an attacker's ability to escape detection by exploiting the NN model's inherent weaknesses significantly diminishes the value of such NN-based detectors.
While previous research \cite{erba2019real} \cite{zizzo2019intrusion} \cite{kravchik2019efficient} studied adversarial evasion attacks on NN ICS detectors, to the best of our knowledge, our study is the first to address poisoning attacks, which are particularly relevant in the online training setting common in ICS detectors.

Our results answer both of the research questions posed in Section \ref{sec:introduction}.
We proposed two algorithms for generating poisoning sequences for multivariate time series data, given the target attack and a trained model.
Both algorithms were found to be effective and capable of producing the desired effect on the model under attack.
The interpolative algorithm is faster, but is in most of the cases less potent than the back-gradient optimization algorithm, which was able to generate poisoning data for attacks with greater magnitude.
The results also confirmed the effectiveness of the initial poison choice algorithm, as it led to shorter poisoning sequences.

After validating the algorithms' effectiveness at generating poisoning sequences, our experiments with simple synthetic data revealed autoencoders' resilience to such poisoning.
Both algorithms were unable to produce arbitrary attacks; the successful poisoning was limited to moderate magnitudes and certain attack locations.
For some attack locations it was not possible to find poisoning for any significant attack without triggering an alert either on the attack or the valid input.
We demonstrated the near linear growth of the required amount of poisoning points with the increasing size of training input.
This finding points at an additional impediment the attacker will need to cope with.

The application of the proposed algorithms to the real data from the SWaT dataset produced the most encouraging results.
While we were able to confirm the algorithms' ability to poison the trained model for the minor attacks, in the vast majority of cases, the out-of-the-box model was already robust to poisoning attempts.
Furthermore, adding more features to the model increased its robustness, delivering valuable confirmation of autoencoder-based ICS attack detectors' trustworthiness.

In a broader context, our results demonstrate that not all NN-based systems are equally vulnerable to adversarial attacks.
For data with a strong internal structure, as is the case of the physics-bound characteristics of a real-world process, NN models appear to be very challenging to manipulate adversarially.
Another possible reason for our models' robustness is their small number of parameters.
While this small amount was sufficient for capturing the internal data dependencies, it was not possible to tweak the parameters without influencing the results the model produces on untainted data.

Some limitations and future directions of this study are worth noting.
First, this study only evaluated autoencoder-based detectors.
However, the  poisoning algorithms proposed are agnostic to the detector's architecture; therefore, studying the robustness of other NN architectures is a topic for future research.
Another important issue for future studies is increasing the efficiency of the poisoning algorithms.
In their current form, they require a very long time to calculate, thus making them inappropriate for real-time poisoning.
Another direction worth exploring is the application of the proposed algorithms to other learning optimizers, in addition to the gradient descent optimizer covered in this study.
According to \cite{maclaurin2015gradient}, other popular optimizers, such as RMSProp \cite{tieleman2012lecture} and Adam \cite{kingma2014adam}, can be traced backwards and thus should be appropriate for our back-gradient optimization poisoning.
In addition, future research will explore the transferability of poisoning between models trained with different optimizers.
Lastly, in this study we used a data-only approach for the algorithms' verification.
Validating the findings in a real-world testbed would be a natural next stage of this research, possibly after finding a way to accelerate poisoning sequence generation.

To conclude, in this study, we proposed and validated two algorithms for poisoning online-trained NN regression-based detectors for multivariate time series data.
We found that autoencoder-based detectors are robust to such poisoning attacks.
Beyond ICS attack detection, these algorithms can be applied in other areas, such as signal processing, speech processing, and medical diagnostics.
Examining the robustness of NNs to poisoning in these domains will be of great benefit to society.

\section{Acknowledgment}
The authors thank iTrust Centre for Research in Cyber Security, Singapore University of Technology and Design for creating and providing the SWaT and WADI datasets, and Ishai Rosenberg for his valuable insights.

\bibliographystyle{IEEEtran}
\bibliography{refs} 

\end{document}